\documentclass[conference]{IEEEtran}
\ifCLASSINFOpdf
\else
\fi
\providecommand{\keywords}[1]
{
  \small	
  \textbf{\textit{Keywords---}} #1
}

\usepackage{hyperref}
\hypersetup{
    colorlinks=true,
    linkcolor=blue,
    filecolor=blue,      
    urlcolor=blue,
    pdftitle={PyGAD Paper},
    pdfpagemode=FullScreen}

\usepackage{graphicx}

\usepackage{xcolor}
\usepackage{xparse}
\usepackage{listings}
\NewDocumentCommand{\codeword}{v}{%
\texttt{\textcolor{blue}{#1}}%
}

\usepackage{amsmath}

\usepackage{enumitem}

\definecolor{verde}{rgb}{0.25,0.5,0.35}
\definecolor{jpurple}{rgb}{0.5,0,0.35}
\definecolor{darkgreen}{rgb}{0.0, 0.2, 0.13}
\usepackage{listings}

\newcommand{\estiloPython}{
  \lstset{ %
    language=Python,                
    basicstyle=\footnotesize,       
    numbers=left,                   
    numberstyle=\tiny\color{gray},  
    stepnumber=1,                   
    linewidth=8.3cm,
    numbersep=5pt,                  
    backgroundcolor=\color{white},  
    showspaces=false,               
    showstringspaces=false,         
    showtabs=false,                 
    frame=single,                   
    rulecolor=\color{black},        
    tabsize=2,                      
    captionpos=b,                   
    breaklines=true,                
    breakatwhitespace=false,        
    title=\lstname,                 
    keywordstyle=\color{blue},      
    commentstyle=\color{darkgreen},   
    stringstyle=\color{red},      
    escapeinside={\%*}{*)},         
    morekeywords={*,pygad,...}          
}}

\title{Estilos para escrever Código Fonte em \LaTeX}
\author{Taciano}

\IEEEoverridecommandlockouts
\usepackage{tikz}
\usepackage{textcomp}
\usepackage{lipsum}
\usepackage{hyperref}

\newcommand\copyrighttext{
  \footnotesize \textcopyright 2021 Ahmed Fawzy Gad. Published at arXiv on June, 10 2021.
  Donation is open at PayPal (\href{https://www.paypal.com/paypalme/ahmedfgad}{paypal.me/ahmedfgad} or e-mail \href{mailto:ahmed.f.gad@gmail.com}{ahmed.f.gad@gmail.com}), OpenCollective (\href{https://opencollective.com/pygad}{opencollective.com/pygad}), and Interac e-Transfer (\href{mailto:ahmed.f.gad@gmail.com}{ahmed.f.gad@gmail.com}).}
\newcommand\copyrightnotice{
\begin{tikzpicture}[remember picture,overlay]
\node[anchor=south,yshift=10pt] at (current page.south) {\fbox{\parbox{\dimexpr\textwidth-\fboxsep-\fboxrule\relax}{\copyrighttext}}};
\end{tikzpicture}%
}

\begin{document}
%
\title{PyGAD: An Intuitive Genetic Algorithm Python Library}

\author{\IEEEauthorblockN{Ahmed Fawzy Gad}
\IEEEauthorblockA{School of Electrical Engineering and Computer Science\\
University of Ottawa\\
Ottawa, ON, Canada\\
\href{mailto:agad069@uottawa.ca}{agad069@uottawa.ca}}}


\maketitle

\copyrightnotice

\begin{abstract}
This paper introduces PyGAD, an open-source easy-to-use Python library for building the genetic algorithm. PyGAD supports a wide range of parameters to give the user control over everything in its life cycle. This includes, but is not limited to, population, gene value range, gene data type, parent selection, crossover, and mutation. PyGAD is designed as a general-purpose optimization library that allows the user to customize the fitness function. Its usage consists of 3 main steps: build the fitness function, create an instance of the pygad.GA class, and calling the pygad.GA.run() method. The library supports training deep learning models created either with PyGAD itself or with frameworks like Keras and PyTorch. Given its stable state, PyGAD is also in active development to respond to the user's requested features and enhancement received on GitHub \href{https://github.com/ahmedfgad/GeneticAlgorithmPython}{github.com/ahmedfgad/GeneticAlgorithmPython}. PyGAD comes with documentation \href{https://pygad.readthedocs.io}{pygad.readthedocs.io} for further details and examples.
\end{abstract}



%
\IEEEpeerreviewmaketitle

\section{Introduction}
Nature has inspired computer scientists to create algorithms for solving computational problems. These naturally-inspired algorithms are called evolutionary algorithms (EAs) \cite{SimonEA} where the initial solution (or individual) to a given problem is evolved across multiple iterations aiming to increase its quality. The EAs can be categorized by different factors like the number of solutions used. Some EAs evolve a single initial solution (e.g. hill-climbing) and others evolve a population of solutions (e.g. particle swarm or genetic algorithm).

The genetic algorithm (GA) a biologically-inspired EA that solves optimization problems inspired by Darwin's theory ``survival of the fittest'' \cite{SimonEA, GadGA}. Originally, the GA was not designed for being a computational algorithm rather understanding how natural selection happens. Later, it becomes one of the most popular computational EAs.

The core concepts of the GA are derived from the natural solution. There are several organisms called population. The members of the population may have the ability to mate and reproduce new organisms. Because the good organisms can produce better offspring, then the good members of the population are selected for mating and others die. When 2 good organisms mate, their offspring will likely be better specially when some new, hopefully, better, changes are introduced. These concepts form the core of the GA in computer science.

\begin{figure}[t]
    \centering
    \includegraphics[width=8.7cm]{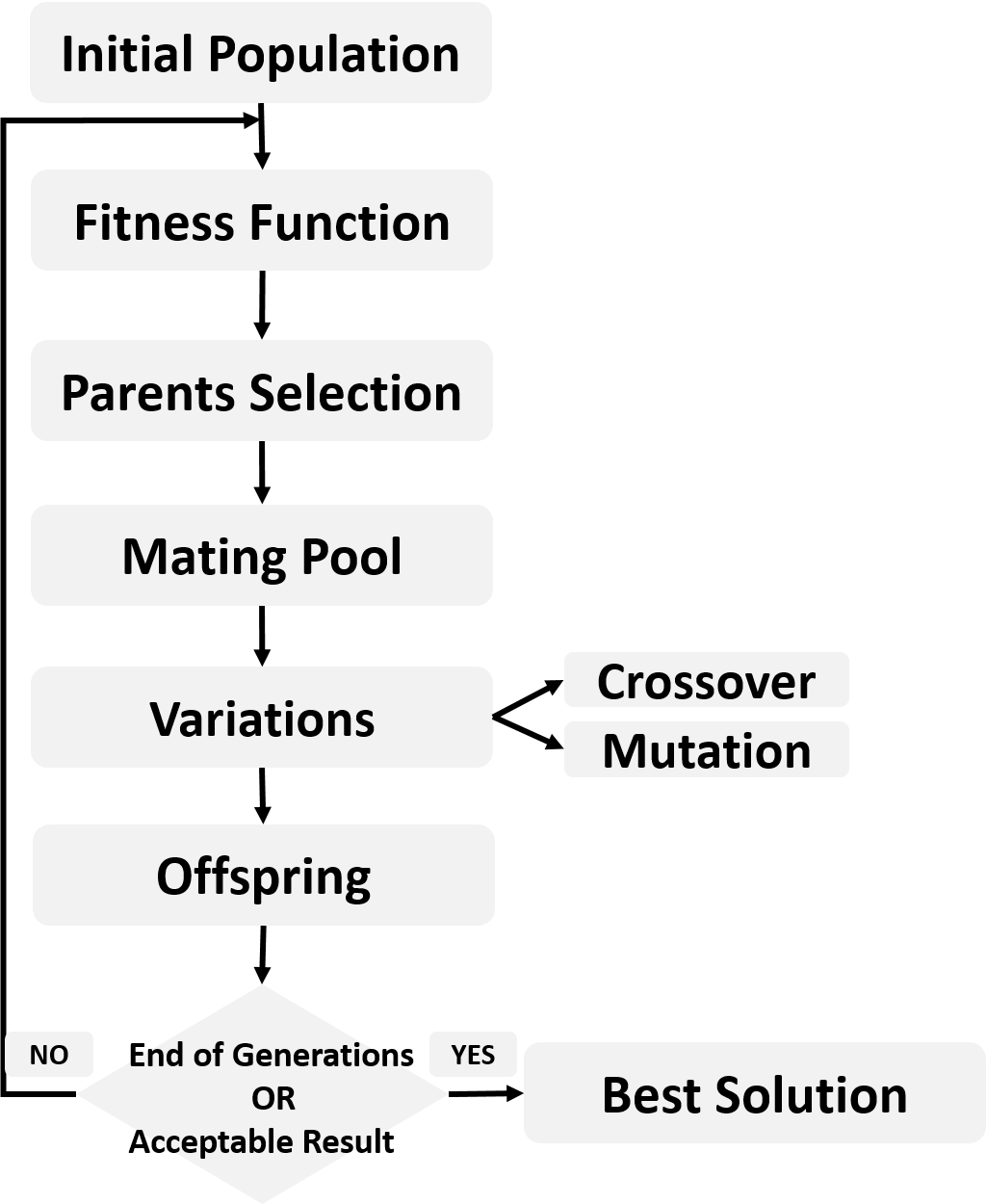}
    \caption{Flowchart of the genetic algorithm.}
    \label{fig:gasteps}
\end{figure}

The GA steps are explained in Figure \ref{fig:gasteps} \cite{GadGA}. The first step creates an initial population of solutions for the problems. These solutions are not the best for the problem and thus evolved. The evolution starts by selecting the fittest solutions as parents based on a maximization fitness function. Each pair of parents mate to produce one or more children. Each child shares genes from its 2 parents using the crossover operation. 

To go beyond the parents' capabilities, the mutation is introduced to make some changes to the children. Due to the randomness in these changes, some children may be fitter or worse than their parents. By mating the best parents and producing more children, a better solution is likely found after each generation. Either the offspring only or combined with the parents form the next generation's population. The process is repeated for a limited number of generations or until a satisfactory solution is found where no further optimization is necessary. 

There are many parameters to tune in to make the GA fits the problem. For experimentation, it is essential to use an easy tool for building the genetic algorithm. 

This paper introduces PyGAD, an open-source intuitive Python library for optimization using the genetic algorithm. PyGAD was released in April 2020 and has over 185K installations at the time of writing this paper. The library supports single-objective optimization with a wide range of parameters to customize the GA for different types of problems in an easy-to-use way with less effort. PyGAD has a lifecycle to trace how everything is working from population creation until finding the best solution. The lifecycle can also be customized to help researchers alter its sequence, enable or disable some operations, make modifications, or introduce new operators. PyGAD works with both decimal and binary representations. The genes can be of \texttt{int}, \texttt{float}, or any supported \texttt{NumPy} numeric data types (\texttt{int}/\texttt{uint}/\texttt{float}). Given the advantages of the GA over the gradient-based optimizers, PyGAD supports training Keras and PyTorch models.

The paper is organized so that section \ref{relatedwork} covers the related work, section \ref{proposedlibrary} extensively introduces PyGAD and briefly compares PyGAD with DEAP and LEAP, and finally, section \ref{conclusion} draws conclusions. Appendix \ref{AppendixA} lists some resources to know more about PyGAD and Appendix \ref{AppendixB} lists the GitHub links of some projects built with PyGAD.

\section{Related Work}
\label{relatedwork}

There are already existing Python libraries for building the genetic algorithm. Some examples include:

\begin{enumerate}[label=\Alph*]
  \item \href{https://deap.readthedocs.io}{DEAP (Distributed Evolutionary Algorithms in Python)} \ref{deap}
  \item \href{http://pyevolve.sourceforge.net}{Pyevolve} \ref{pyevolve}
  \item \href{https://github.com/danielwilczak101/EasyGA}{EasyGA} \ref{easyga}
  \item \href{https://leap-gmu.readthedocs.io/en/latest}{LEAP (Library for Evolutionary Algorithms in Python)} \ref{leap}
\end{enumerate}

This section gives an overview of these libraries by explaining their objectives and limitations.

\subsection{DEAP}
\label{deap}
\href{https://deap.readthedocs.io}{DEAP (Distributed Evolutionary Algorithms in Python)} \cite{DEAP} is considered one of the most common Python libraries for optimization using the genetic algorithm based on the number of installations, GitHub issues, and stars (4.2K). One of the reasons is being one of the first libraries about EAs which was published in 2012. DEAP supports other algorithms than GA like non-dominated sorting genetic algorithm II (NSGA-II), particle swarm optimization (PSO), and evolution strategy (ES). Its latest release at Python Package Index (PyPI) is 1.3.1 released on Jan 2020.

DEAP uses 2 main structures which are \texttt{creator} and \texttt{toolbox}. The \texttt{creator} module is used for creating data types like fitness, individual, and population. These data types are empty and filled using the \texttt{toolbox} module. Given the special structure of DEAP, the user would take some time until understanding getting familiar. It needs more than a beginner's level in Python.

One problem about DEAP is being not user-friendly as the user has to do some efforts for each optimization problem. For example, the user has to write extra code for creating the population and filling it with appropriate values. The motivation for that is to enable the user to create custom data types that are not supported. But this boilerplate code unnecessarily increases the length of the Python script and makes it uncomfortable for users to write and understand more syntax about DEAP. Sometimes, the scripts are hard to read.

Moreover, the users have to create a main function in which everything is grouped in an evolutionary loop. The loop in this user-defined function should is where the user needs to follow the GA pipeline by 1) calculating the fitness function, 2) selecting the parents, 3) applying crossover and mutation, 4) and repeating that for several generations. This is not the best interface for users who try to focus on the experiments and save time on the other stuff. Writing the evolutionary loop makes the library non-friendly.

Even if the library supports some ready-to-use algorithms that save time building the main function, each algorithm does a specific task and is limited in its features. With the few parameters each algorithm accepts, there is little customization possible. Examples of these algorithms are \texttt{eaSimple}, \texttt{eaMuPlusLambda}, and \texttt{varOr}.

The library leaves much stuff to be built by the end-user which makes the user feel lost between the modules, classes, and functions needed to customize the library to solve a problem. For example, building a population of mixed data types requires the user to:

\begin{enumerate}
  \item Register a data type for each gene.
  \item Register an individual that combines those gene data types.
  \item Register a population that uses that individual.
  \item Build the population
\end{enumerate}

There is no way to restrict the gene values to a set of sparse discrete values. This is necessary for some problems where the gene value may not fall within a regular range. To select which genes to mutate, DEAP only supports the mutation probability. There is no way to specify the exact number of genes to be mutated. 

DEAP only supports the traditional mutation operators where all solutions, regardless of their fitness value, are given equal mutation probability. This would distort the high-quality solutions.

Another major drawback of DEAP is the lack of means of visualizing the results after the evolution completes \cite{DEAPReview}. The users have to manually create their plots to summarize the results.

\subsection{Pyevolve}
\label{pyevolve}
\href{http://pyevolve.sourceforge.net}{Pyevolve} is another pure Python library for building the genetic algorithm \cite{Pyevolve}. Even that it is published in 2009, it is less popular than DEAP and this is based on the total number of installations (50K for all the time), GitHub stars (301), and citations. It also has a limited community compared to DEAP. 

Its scripts start by creating the fitness function, preparing the chromosome, setting some parameters like which operators to use, create an instance of the \texttt{GSimpleGA} class and then call the \texttt{evolve()} method to start the evolution.

Like DEAP, this library has boilerplate code for configurations that makes its scripts unnecessarily longer even for simple problems. The supported mutation operators apply mutation equally to all solutions independent of their fitness value.

Even that Pyevolve is easier than DEAP, but it is limited in its features. It supports some predefined data types for the genes like integer and real. To create new data types, then more details about some classes in the library are needed like \texttt{GenomeBase.GenomeBase} which may be problematic for some users.

A comparison between DEAP and Pyevolve shows that the number of code lines needed to solve the OneMax problem is 59 for DEAP and 378 for Pyevolve \cite{DEAP}. Given the simplicity of the problem, Pyevolve needed too many lines.

This library is no longer maintained as the latest version was released at the end of 2014 and the most recent commit on its GitHub project was at the end of 2015.

\subsection{EasyGA}
\label{easyga}
The \href{https://github.com/danielwilczak101/EasyGA}{EasyGA} library allows only defining a continuous range for the gene values. If the range has some exclusions or if the gene values do not follow a range at all, then there is no way to define the gene values. Moreover, the user has to build the solutions manually without given a simple interface. Commonly, users would like to focus more on the algorithm itself and save time building additional modules specially if they are not involved in Python.

This library has a limited number of operators for crossover, mutation, and parent selection. Given the 

The EasyGA library supports a random mutation operation that applies mutation over any solution in the population, including parents, and is not restricted to the new offspring. It randomly selects the solutions to mutate. This is against the nature of the GA as only random changes could be introduced to the offspring, not the parents.

The users have to know about decorators, at some stage, to build their operators. While writing the paper, the GitHub project of EasyGA has only 22 stars.

\subsection{LEAP}
\label{leap}
\href{https://leap-gmu.readthedocs.io/en/latest}{LEAP (Library for Evolutionary Algorithms in Python)} is another recent Python library published in 2020 for EAs that supports the genetic algorithm \cite{LEAP}. This library has 3 core classes which are \texttt{Individual}, \texttt{Decoder}, and \texttt{Problem}. The decoder is responsible for converting the genes from one form to another to calculate the fitness value for each individual given the current problem. 

According to the examples posted by the developers, the decoder is usually set to \texttt{IdentityDecoder()} which means the gene is translated to itself. This is a design issue in the library. Maybe this feature is helpful in some specific problems but the library uses it as something essential for all types of problems.  It is better to work directly on the genes without having to decode them to another form or leave that decoding part to the fitness function. 

Even it is published in July 2020, the library has many missing features as mentioned in its documentation. One of these missing features is the mixed data representation in the individual. There is no lifecycle in LEAP to help trace what is happening in each generation.

Even if one of the objectives of LEAP is to make it simpler than the other libraries, the user still has to write the evolution loop. This results in more lines to solve a problem. Moreover, the user has to take care of calling a function that increases the generations counter by calling the \texttt{util.inc\_generation()} function. Avoiding to call it causes an infinite evolution loop. This would be a problem for users with less experience. 

Creating and managing the evolution loop is against another objective of LEAP to make it suitable for all types of software users (users who solve problems with existing tools).  

This library is not popular as it has a total of 3.4K installations since publication in addition to just 39 starts in the GitHub project. 

\section{Proposed PyGAD Library}
\label{proposedlibrary}

\href{https://pygad.readthedocs.io}{PyGAD} is an open-source Python library for optimization using the genetic algorithm. It is published in April 2020 at PyPI (\href{https://pypi.org/project/pygad}{pypi.org/project/pygad}). Its GitHub project has over 525 stars (\href{https://github.com/ahmedfgad/GeneticAlgorithmPython}{github.com/ahmedfgad/GeneticAlgorithmPython}). With over 185K installations over 1 year, PyGAD is the most rising library compared to the other libraries. 

The name PyGAD has 3 parts:
\begin{enumerate}
  \item \textbf{\textit{Py}} which means it is a \textit{Python} library. This is a common naming convention for Python libraries.
  \item \textbf{\textit{GA}} stands for \textit{genetic algorithm}.
  \item \textbf{\textit{D}} for \textit{decimal} because the library was originally supporting only decimal genetic algorithm. Now, it supports both decimal and binary genetic algorithm.
\end{enumerate}

This section gives an overview of \href{https://pygad.readthedocs.io}{PyGAD} \ref{pygadoverview}, discusses the steps of its usage \ref{pygadusage}, its lifecycle \ref{pygadlifecycle}, and main features in PyGAD \ref{pygadmainfeatures}. 

\subsection{PyGAD Overview}
\label{pygadoverview}

PyGAD is an intuitive library for optimization using the genetic algorithm. It is designed with 2 main objectives:
\begin{enumerate}
    \item Making everything as simple as possible for the users with the least knowledge.
    \item Giving the user control over everything possible.
\end{enumerate}

The simplicity comes from using descriptive names for the classes, methods, attributes, and parameters. This is in addition to making things straightforward to build the genetic algorithm and specify a wide range of easy-to-understand configuration parameters. There are fewer classes, methods, or functions to call to solve a problem compared to the other libraries. 

The users do not have to build the evolution loop in any situation as PyGAD supports an elastic lifecycle that can be altered. This includes, but is not limited to, enabling or disabling the mutation or crossover operators and overriding them to build new operators for research purposes.

The 7 modules included in PyGAD are:
\begin{enumerate}
    \item \texttt{pygad.pygad}: It is the main module which builds everything in the genetic algorithm. This module is implicitly imported when the library itself is imported.
    \item \texttt{pygad.nn}: Builds fully-connected neural networks (FCNNs) from scratch using only NumPy.
    \item \texttt{pygad.gann}: Uses the \texttt{pygad} module to train networks build using the \texttt{nn} module.
    \item \texttt{pygad.cnn}: Similar to the \texttt{nn} module but builds convolutional neural networks (CNNs).
    \item \texttt{pygad.gacnn}: Similar to the \texttt{gann} module but trains CNNs.
    \item \texttt{pygad.kerasga}: Trains Keras models using the \texttt{pygad} module.
    \item \texttt{pygad.torchga}: Trains PyTorch models using the \texttt{pygad} module.
\end{enumerate}

Given that the \texttt{pygad.pygad} is the most critical module in the library, it is given the most attention.

PyGAD has detailed documentation that covers all of its features with examples. Moreover, the source code of various projects built using PyGAD is explained in tutorials. A list of some of these tutorials is available in Appendix \ref{AppendixA}.

PyGAD gained popularity in the last months and some of its English articles and tutorials are translated to different languages like \href{https://pygad.readthedocs.io/en/latest/Footer.html#korean}{Korean}, \href{https://pygad.readthedocs.io/en/latest/Footer.html#turkish}{Turkish}, \href{https://pygad.readthedocs.io/en/latest/Footer.html#hungarian}{Hungarian}, \href{https://pygad.readthedocs.io/en/latest/Footer.html#chinese}{Chinese}, and \href{https://pygad.readthedocs.io/en/latest/Footer.html#russian}{Russian}. A list of such translated articles and tutorials is found in the \href{https://pygad.readthedocs.io/en/latest/Footer.html#pygad-in-other-languages}{PyGAD in Other Languages} section of the documentation.

The documentation of PyGAD has a section called \href{https://pygad.readthedocs.io/en/latest/Footer.html#release-history}{Release History} to summarize the changes and additions in each release.

Appendix \ref{AppendixA} has extra reading resources about PyGAD. Appendix \ref{AppendixB} lists some projects built with PyGAD.

\subsection{PyGAD Usage}
\label{pygadusage}

There are 3 core steps to use PyGAD:
\begin{enumerate}
    \item Build the fitness function.
    \item Instantiate the \texttt{pygad.GA} class with the appropriate configuration parameters.
    \item Call the \texttt{run()} method to start the evolution.
\end{enumerate}

For the following equation with 3 inputs, we can use PyGAD to find the values of $w_1$, $w_2$, and $w_3$ that satisfy equation:
\begin{gather*}
    Y = w_1X_1 + w_3X_3 + w_3X_3 \\
    Where \: Y=44, \; X_1=4, \; X_2=-2, \; and \; X_3=3.5
\end{gather*}

A basic PyGAD example that solves this problem is given in Listing \ref{lst:pygadexample}. Line \ref{line:importpygad} imports the library. This import statement implicitly imports the \texttt{pygad} module. The \texttt{NumPy} library is also imported in Line \ref{line:importnumpy} because it is used in the fitness function.

Line \ref{line:equationin} creates a Python list to hold the 3 inputs and line \ref{line:equationout} holds the output.

The fitness function in PyGAD is a regular Python function that accepts 2 parameters:
\begin{enumerate}
    \item The solution evolved by the genetic algorithm as a 1D vector. The function should return a single numeric value representing the fitness of this solution.
    \item The index of the solution within the population.
\end{enumerate}

The length of the solution in this example is 3, one value for each weight in the equation. The fitness function must be a maximization function (the higher the fitness value the better the solution).

The fitness function is defined from line \ref{line:fitnessfunctionstart} to line \ref{line:fitnessfunctionend} in Listing \ref{lst:pygadexample}. The function calculates the sum of products between each value in the solution and its corresponding input. The absolute difference is calculated between the sum and the output. 

Returning the absolute difference makes it a minimization function. This is why the result is returned as $1.0/abs$. A tiny value of $0.000001$ is added to the denominator to avoid diving by zero.

A new instance of the \texttt{pygad.GA} class is created in line \ref{line:pygadgainstance}. All configuration parameters are grouped in the \texttt{pygad.GA} class's constructor. With the available IDE's, the user can easily check the names of all available parameters. This way the user does not have to memorize the names of functions or classes compared to the other libraries. 

There are 5 required parameters that must be specified for each problem:
\begin{enumerate}
    \item \texttt{num\_generations}: The number of generations/iterations.
    \item \texttt{sol\_per\_pop}: The number of solutions/chromosomes/individuals in the population (i.e. population size).
    \item \texttt{num\_parents\_mating}: The number of solutions to be selected from the population as parents for mating and producing the offspring.
    \item \texttt{num\_genes}: The number of genes in each solution.
    \item \texttt{fitness\_func}: The fitness function.
\end{enumerate}

These are the minimum parameters to use PyGAD. Note that the names of the parameters are self-describing. The documentation has information about the classes, parameters, attributes, methods, and functions in all PyGAD modules. \hfill \break

\begin{scriptsize}
\estiloPython
\begin{minipage}{\linewidth}
\begin{lstlisting}[caption={PyGAD example.}, label=lst:pygadexample, escapechar=|]
import pygad |\label{line:importpygad}|
import numpy |\label{line:importnumpy}|

equation_inputs = [4, -2, 3.5] |\label{line:equationin}|
Y = 44 |\label{line:equationout}|

def fitness_func(solution, solution_idx): |\label{line:fitnessfunctionstart}|
    out = numpy.sum(solution * equation_inputs)
    fitness = 1.0/(numpy.abs(out - Y)+0.000001)
    return fitness |\label{line:fitnessfunctionend}|

ga_instance = pygad.GA(num_generations=100, |\label{line:pygadgainstance}|
                       sol_per_pop=10,
                       num_parents_mating=5,
                       num_genes=3,
                       fitness_func=fitness_func)

ga_instance.run() |\label{line:runga}|
\end{lstlisting}
\end{minipage}
\end{scriptsize}

In Listing \ref{lst:pygadexample}, a population of size 10 is evolved through 100 generations. Out of the population, 5 solutions are selected for mating. Each solution has 3 genes where the fitness value is calculated using the Python function named \texttt{fitness\_func}. 

The instances of the \texttt{pygad.GA} class has a method called \texttt{run()} which runs the genetic algorithm to start evolving the initial population according to the selected parameters. This is the minimal code for optimizing this problem using PyGAD.

Once the \texttt{run()} method completes, additional methods can be called to find information about the solution found by PyGAD. Two of these methods are:
\begin{enumerate}
    \item \texttt{best\_solution()}: Returns the following information about the best solution found by PyGAD:
    \begin{enumerate}
        \item The parameters of the best solution (e.g. the 3 weights for the problem solved in Listing \ref{lst:pygadexample}.
        \item The fitness value of the best solution.
        \item The index of this solution in its population.
    \end{enumerate}
    \item \texttt{plot\_result()}: Creates a plot showing how the fitness value evolves by each generation. This method returns the figure in case the user would like to save it.
\end{enumerate}

\begin{figure}[t]
    \centering
    \includegraphics[width=8.7cm]{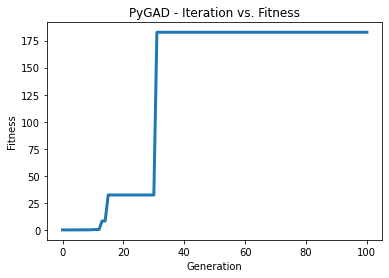}
    \caption{Evolution of the fitness value for 100 generations.}
    \label{fig:pygadexample_plot}
\end{figure}

These 2 methods are called in Listing \ref{lst:pygadexampleplot}. Figure \ref{fig:pygadexample_plot} shows that the best solution is found after 31 generations only with a fitness value of 182.698. The best solution's parameters can be plugged into the equation to return the predicted value. \hfill \break

\begin{scriptsize}
\estiloPython
\begin{minipage}{\linewidth}
\begin{lstlisting}[caption={Evolution results.}, label=lst:pygadexampleplot, escapechar=|]
solution, solution_fitness, solution_idx = ga_instance.best_solution()
fig = ga_instance.plot_result()
\end{lstlisting}
\end{minipage}
\end{scriptsize}

For the OneMax optimization problem, it is solved with PyGAD in just 15 lines of code compared to 45 for DEAP and 34 for LEAP.

For 3 different runs with 100 generations each, the average time for PyGAD is 0.14 seconds compared to 0.65 for DEAP and 0.052 for LEAP. The trouble with LEAP is that the optimal solution was not found even after 1,000 generations. The maximum number of ones did not even reach $90/100$ after tens of trials with the code published by the developer at GitHub. 

\subsection{PyGAD Lifecycle}
\label{pygadlifecycle}

\begin{figure}[t]
    \centering
    \includegraphics[width=8.7cm]{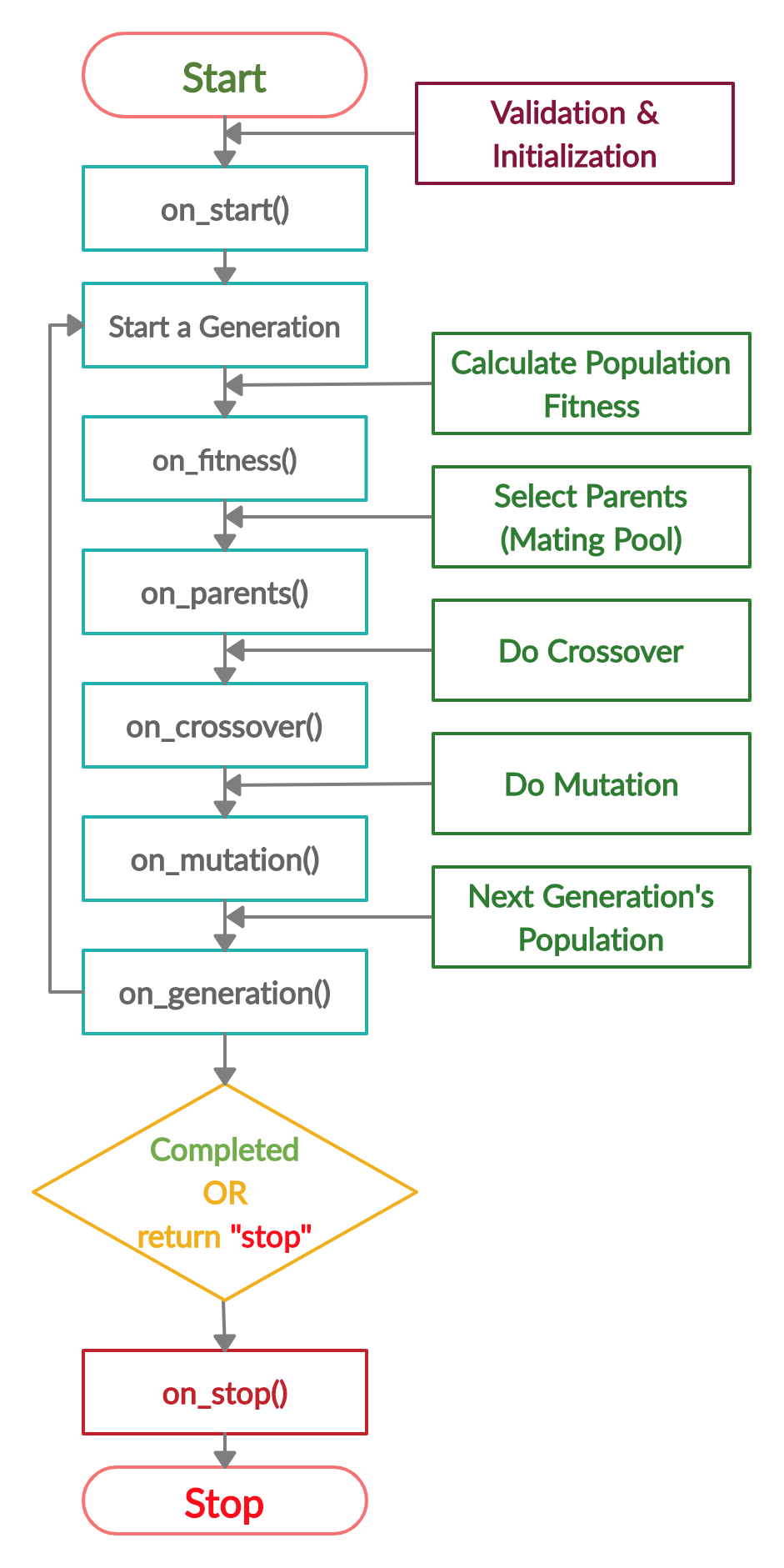}
    \caption{PyGAD lifecycle.}
    \label{fig:pygad_lifecycle}
\end{figure}

PyGAD has a lifecycle that helps to user to keep track and control all different stages of the evolution process. Figure \ref{fig:pygad_lifecycle} shows the PyGAD lifecycle. Note that the side blocks refer to operations done between one state and another.

The passed parameters to the \texttt{pygad.GA} class's constructor are validated. Then, instance attributes are initialized according to the passed parameters. Some of these attributes include \texttt{population} which is a NumPy array holding all solutions in the population.

Once an instance of the \texttt{pygad.GA} class is created successfully, then it is time to start the lifecycle of PyGAD by calling the \texttt{run()} method. For the 7 states in PyGAD lifecycle, the following 7 callback functions exist:
\begin{enumerate}
    \item \texttt{on\_start()}: Called once after the \texttt{run()} method is called.
    \item \texttt{on\_fitness()}: Called after calculating the population fitness in each generation.
    \item \texttt{on\_parents()}: Called in each generation after the parents are selected.
    \item \texttt{on\_crossover()}: For each generation, it is called after applying the crossover operation.
    \item \texttt{on\_mutation()}: For each generation, it is called after applying the mutation operation.
    \item \texttt{on\_generation()}: Called at the end of each generation.
    \item \texttt{on\_stop()}: Called once after the \texttt{run()} method stops.
\end{enumerate}

The user can assign a Python function, with the appropriate parameters, to any of these callback functions. By doing this, the user can track and control everything from the start to the end of the evolution. This includes, but is not limited to, getting information about the population, fitness, selected parents, crossover or mutation results, and the best solution yet found. The user can also do some pre-processing or post-processing tasks in the \texttt{on\_start()} and \texttt{on\_stop()} functions, respectively. 

For research purposes and supporting an operator that is not supported by PyGAD, the user can define custom crossover and mutation operators in the callback functions. 

The \texttt{on\_generation()} callback function can be used to add some conditions and alter the execution. For example, returning the string "stop" immediately stops the \texttt{run()} method earlier before completing all generations.

\subsection{PyGAD Features}
\label{pygadmainfeatures}

The previous sections should have covered some features of PyGAD. This section highlights the main features supported by PyGAD that make it distinctive compared to the other libraries. For more details, check its documentation \href{https://pygad.readthedocs.io}{pygad.readthedocs.io}.

\begin{enumerate}
  \item PyGAD is very intuitive to use. Its steps are self-explanatory and easy to follow in a user-friendly way. With just 3 simple steps, PyGAD can optimize different types of problems.
  \item All parameters are grouped in the constructor of the \texttt{pygad.GA} class. This helps the user to be focused and not distracted. 
  \item A user-defined gene value space, either for all genes at once or for each gene, can be created using the \texttt{gene\_space} parameter. This is helpful even if the gene's values do not follow a certain pattern. If a gene has values enclosed within a given range, then the gene value can be randomly generated from a user-defined range.
  \item It is possible to control the gene data type, for all genes or each gene, using the \texttt{gene\_type} parameter. The supported data types are Python's \texttt{int} and \texttt{float} in addition to all \texttt{int}/\texttt{uint}/\texttt{float} types in the \texttt{NumPy} library.
  \item Using the \texttt{initial\_population} parameter, a user-defined initial population can be used as the start point. This helps to continue where the evolution stopped. PyGAD builds an initial population randomly if this parameter is not set.
  \item PyGAD has a lifecycle to keep track of everything. A call to a callback function can be triggered for each step in the lifecycle.
  \item The user can stop the evolution at any time. For example, returning the string ``stop'' from the \texttt{on\_generation()} callback function stops PyGAD at the current generation.
  \item It is easy to access the best solutions across all generations in addition to their fitness using the \texttt{best\_solutions} and \texttt{best\_solutions\_fitness} attributes, respectively.
  \item The crossover and mutation operations can be disabled by setting \texttt{crossover\_type} or \texttt{mutation\_type} to \texttt{None} and then plug a user-defined operation in the lifecycle.
  \item There are different ways to specify the number of genes to mutate. According to the user's preference, this can be specified as probability (\texttt{mutation\_probability}), percentage (\texttt{mutation\_percent\_genes}), or an explicit number of genes (\texttt{mutation\_num\_genes}).
  \item PyGAD supports adaptive mutation so that the user controls how mutation is applied based on the solution's fitness. For the high-quality solutions, low mutation probability/percentage/number is expected compared to low-quality solutions. This helps to maintain the quality of the good solutions in addition to increasing the quality of the bad solutions.
  \item The \texttt{mutation\_by\_replacement} \texttt{bool} parameter selects whether the mutation adds to or replaces the gene value.
  \item A \texttt{bool} parameter called \texttt{allow\_duplicate\_genes} helps to accept or reject duplicate values in the same chromosome. For this to work, there must be enough value space to guarantee unique genes' values.
  \item Ability to control the number of parents to keep in the next generation using the \texttt{keep\_parents} parameter. This helps to save 0 or more parents to keep the fitness curve increasing.
  \item The instance attributes in the \texttt{pygad.GA} class starting with \texttt{last\_generation\_} help to keep track of the outcomes of each generation.
  \item PyGAD supports training \href{https://keras.io}{Keras} and \href{https://pytorch.org}{PyTorch} models using the \texttt{pygad.kerasga} and \texttt{torchga} modules, respectively.
  \item PyGAD also has its own modules to build and train neural networks (\texttt{pygad.nn}, \texttt{pygad.gann}, \texttt{pygad.cnn}, and \texttt{pygad.gacnn}).
  \item The interface to use any PyGAD feature is very simple and does not require much knowledge about Python compared to the other libraries. The minimum level is to know how to create variables, how to define a Python function that accepts parameters and returns something, little about classes like creating instances in addition to accessing class/instance attributes.
  \item PyGAD has built-in support to visualize the results. For example, the \texttt{best\_solution()} method shows how the fitness changes by generation.
  \item There are many resources to help you get started with PyGAD. This includes documentation, blog posts, examples, projects, and a community over \href{https://github.com/ahmedfgad/GeneticAlgorithmPython}{GitHub}, \href{https://www.reddit.com/search/?q=PyGAD&sort=relevance}{reddit}, \href{https://stackoverflow.com/search?q=PyGAD}{StackOverflow},  \href{https://www.facebook.com/pygad}{Facebook}, and \href{https://twitter.com/PyGADLib}{Twitter}.
\end{enumerate}

\section{Conclusion}
\label{conclusion}

This paper proposed a new Python library called PyGAD for single-objective optimization using the genetic algorithm. PyGAD is an intuitive library that makes it easy to optimize problems in just 3 steps: fitness function creation, instantiating the \texttt{pygad.GA} class, and calling \texttt{run()} method. There is a wide range of parameters and attributes to have a high degree of customizing the genetic algorithm. This includes defining a space of values for each gene, customizing each gene's data type, training Keras and PyTorch models, rejecting duplicates, and more. PyGAD has a lifecycle to keep track of the evolution process from the beginning to the end. PyGAD supports a simpler interface for users with less experience with Python to solve problems with few lines of code and even faster than the other libraries.

\section*{Acknowledgment}

I would like to thank everyone who used or showed interest in PyGAD. Some of those people who reported issues or suggested useful features are \href{https://github.com/tfarrag2000}{Tamer Farrag} (Assistant Professor, Misr Higher Institute for Engineering and Technology, Egypt), \href{https://www.linkedin.com/in/hamadakassem}{Hamada Kassem} (RA/TA, Faculty of Engineering, Alexandria University, Egypt), \underline{Curt McDowell},  \href{https://www.mpibpc.mpg.de/17254711/andrei-rozanski}{Andrei Rozanski} (PhD Bioinformatics Specialist, Max Planck Institute for Biophysical Chemistry, Germany), \href{https://www.researchgate.net/profile/Marios-Giouvanakis}{Marios Giouvanakis} (PhD candidate in Electrical \& Computer Engineer, Aristotle University of Thessaloniki, Greece), \href{https://www.linkedin.com/in/l\%C3\%A1szl\%C3\%B3-fazekas-2429a912}{László Fazekas} (CTO Senior Software Developer at Pressenger Ltd, Hungary), and special thanks to \href{https://www.linkedin.com/in/rainer-matthias-engel-5ba47a9}{Rainer Engel} (Imaging Artist and Pipeline Developer, Germany) for his generous suggestions and time offered in inspecting PyGAD. 



%

\begin{appendices}

\section{PyGAD Supplemental Resources}
\label{AppendixA}

This appendix lists some tutorials and articles to get started with PyGAD. 

\begin{itemize}
    \item \href{https://blog.paperspace.com/genetic-algorithm-applications-using-pygad}{5 Genetic Algorithm Applications Using PyGAD}, June 2020, \href{https://www.linkedin.com/in/ahmedfgad}{Ahmed Gad}
    \item \href{https://blog.paperspace.com/building-agent-for-cointex-using-genetic-algorithm}{Building a Game-Playing Agent for CoinTex Using PyGAD}, July 2020, \href{https://www.linkedin.com/in/ahmedfgad}{Ahmed Gad}
    \item \href{https://blog.paperspace.com/working-with-different-genetic-algorithm-representations-python}{Working with Different Genetic Algorithm Representations in PyGAD}, September 2020, \href{https://www.linkedin.com/in/ahmedfgad}{Ahmed Gad}
    \item \href{https://heartbeat.fritz.ai/train-neural-networks-using-a-genetic-algorithm-in-python-with-pygad-862905048429?gi=ba58ee6b4bbd}{Train Neural Networks Using a Genetic Algorithm in Python with PyGAD}, September 2020, \href{https://www.linkedin.com/in/fatima-ezzahra-jarmouni-341a6b167/}{Fatima Ezzahra Jarmouni}
    \item \href{https://blog.paperspace.com/train-keras-models-using-genetic-algorithm-with-pygad}{How To Train Keras Models Using the Genetic Algorithm with PyGAD}, December 2020, \href{https://www.linkedin.com/in/ahmedfgad}{Ahmed Gad}
    \item \href{https://markatsmartersig.wordpress.com/2021/01/04/genetic-algorithm-v-gradient-boosting}{Genetic Algorithm V Gradient Boosting}, January 2021, \href{http://www.smartersig.com/mysportsai.php}{Mark Littlewood}
    \item \href{https://blog.paperspace.com/clustering-using-the-genetic-algorithm}{Clustering Using the Genetic Algorithm with PyGAD}, March 2021, \href{https://www.linkedin.com/in/ahmedfgad}{Ahmed Gad}
    \item \href{https://neptune.ai/blog/train-pytorch-models-using-genetic-algorithm-with-pygad}{Train PyTorch Models Using Genetic Algorithm with PyGAD}, March 2021, \href{https://www.linkedin.com/in/ahmedfgad}{Ahmed Gad}
    \item \href{https://hackernoon.com/how-genetic-algorithms-can-compete-with-gradient-descent-and-backprop-9m9t33bq}{How Genetic Algorithms Can Compete with Gradient Descent and Backprop}, March 2021, \href{https://www.linkedin.com/in/l\%C3\%A1szl\%C3\%B3-fazekas-2429a912}{László Fazekas}
    \item \href{https://neptune.ai/blog/adaptive-mutation-in-genetic-algorithm-with-python-examples}{Adaptive Mutation in Genetic Algorithm with PyGAD Examples}, April 2021, \href{https://www.linkedin.com/in/ahmedfgad}{Ahmed Gad}
    \item \href{https://markatsmartersig.wordpress.com/2021/06/05/pygad-v-gbm}{PyGAD v GBM}, June 2021, \href{http://www.smartersig.com/mysportsai.php}{Mark Littlewood}
\end{itemize}

\section{Projects with PyGAD}
\label{AppendixB}

This is a list of some projects built using PyGAD with their source code:

\begin{itemize}
    \item \href{https://github.com/ahmedfgad/CoinTex/tree/master/PlayerGA}{github.com/ahmedfgad/CoinTex/tree/master/PlayerGA}: Play a game called CoinTex.
    \item \href{https://github.com/ahmedfgad/FlappyBirdPyGAD}{github.com/ahmedfgad/FlappyBirdPyGAD}: Play the flappy bird game.
    \item \href{https://github.com/ahmedfgad/8QueensGenetic}{github.com/ahmedfgad/8QueensGenetic}: Solve the 8-queen puzzle.
    \item \href{https://github.com/ahmedfgad/GARI}{github.com/ahmedfgad/GARI}: Reproduce gray and RGB images. 
\end{itemize}

\end{appendices}

\end{document}